\documentclass{article} 
\usepackage{iclr2026_conference,times}

\usepackage{amsmath,amsfonts,bm}

\def\eqref#1{equation~\ref{#1}}

\def\1{\bm{1}}

\DeclareMathAlphabet{\mathsfit}{\encodingdefault}{\sfdefault}{m}{sl}
\SetMathAlphabet{\mathsfit}{bold}{\encodingdefault}{\sfdefault}{bx}{n}

\usepackage{hyperref}
\usepackage{url}
\usepackage{graphicx}
\usepackage{booktabs}
\usepackage[table]{xcolor} 
\usepackage[table]{xcolor} 
\usepackage{booktabs}      
\usepackage{wrapfig}  
\usepackage[table]{xcolor} 
\usepackage[most]{tcolorbox}
\usepackage{algorithm}
\usepackage{algpseudocode}
\usepackage{amsmath}
\usepackage{xcolor}
\usepackage{listings}
\usepackage{xcolor}
\usepackage{amsmath}
\usepackage{multirow}
\usepackage{hyperref}
\usepackage{multirow}
\usepackage{amsfonts}
\usepackage{amssymb}

\lstdefinestyle{pythonICLR}{
    language=Python,
    basicstyle=\ttfamily\footnotesize,
    keywordstyle=\color{blue}\bfseries,
    stringstyle=\color{orange},
    commentstyle=\color{gray}\itshape,
    morekeywords={GraphState, MultiAgentRoboticSystem},
    showstringspaces=false,
    breaklines=true,
    frame=single,
    backgroundcolor=\color{yellow!5},
    rulecolor=\color{gray!50},
    numbers=left,
    numberstyle=\tiny\color{gray},
    xleftmargin=2em,
    framexleftmargin=1.5em
}

\definecolor{mallviColor}{HTML}{E1FBFF}

\title{MALLVi: A Multi-Agent Framework for Integrated Generalized Robotics Manipulation}

\author{
Mehrshad Taji, \hspace*{0.5em}
Arad Mahdinezhad Kashani\rlap{$^*$}, \hspace*{0.5em}
Iman Ahmadi\rlap{$^*$} \\
\textbf{AmirHossein Jadidi, \hspace*{0.5em}
Saina Kashani, \hspace*{0.5em}
Babak Khalaj} \\
Department of Electrical Engineering \\
Sharif University of Technology \\
\texttt{mehrshad.taji@sharif.edu, arad.mnk81@sharif.edu,} \\
\texttt{iman.ahmadi@sharif.edu, amirhossein.jadidi94@sharif.edu,} \\
\texttt{saina.kashani83@sharif.edu, khalaj@sharif.edu}
}
\begin{document}

\maketitle

{\renewcommand{\thefootnote}{\fnsymbol{footnote}}%
\setcounter{footnote}{0}%
\footnotetext{* Equal Contribution}%
\renewcommand{\thefootnote}{\arabic{footnote}}%
\setcounter{footnote}{0}}

\begin{abstract}

Task planning for robotic manipulation tasks using large language models (LLMs)
is a relatively new phenomenon. Previous approaches have relied on training
specialized models, fine-tuning pipeline components, or adapting LLMs with the
setup through prompt tuning. However, many of these approaches
operate in an open-loop manner and lack robust environmental feedback, making them
fragile in dynamic or unstructured settings. We introduce the MALLVi Framework, a
\textbf{M}ulti-\textbf{A}gent \textbf{L}arge \textbf{L}anguage and \textbf{V}ision framework designed to solve robotic
manipulation tasks that explicitly leverages closed-loop,
feedback-driven interaction with the environment. The agents are provided with
an instruction in human language, along with an image of the current environment
state. After thorough investigation and reasoning, MALLVi generates a series of
realizable atomic instructions necessary for a supposed robot manipulator to
complete the task. After extracting and executing low-level actions through the
downstream agents, a Vision-Language Model (VLM) receives environmental feedback
and prompts the framework either to repeat this procedure until success, or to
proceed with the next atomic instruction.
Rather than relying on a single monolithic model, MALLVi
coordinates multiple specialized LLM agents, each responsible for a distinct
component of the manipulation pipeline. Our work shows that with careful prompt
engineering, the integration of four LLM agents (\textbf{Decomposer}, \textbf{Localizer}, \textbf{Thinker},
and \textbf{Reflector}) can autonomously manage all compartments of a manipulation task—
namely, initial perception, object localization, reasoning, and high-level planning. Moreover, the addition of a \textbf{Descriptor} agent can introduce a visual
memory of the initial environment state in the pipeline.
Crucially, the Reflector agent enables targeted error detection
and recovery by evaluating the completion or failure of each subtask and
reactivating only the relevant agent, avoiding costly global replanning. We validate our framework through experiments conducted both in simulated
environments using VIMABench and RLBench, and in real-world settings. Our
framework handles diverse tasks, from standard manipulation benchmarks to custom
user instructions. Our results demonstrate that iterative,
closed-loop communication among agents significantly improves generalization and
increases average success rates in zero-shot manipulation scenarios.

\end{abstract}

\section{Introduction}
Natural language tasks are rich, contextual, and often complicated —a simple sentence may be broken down to several smaller subtasks. With the advent of large language models (LLMs), attempts in the task-planning field for robotic manipulation have shifted to using complex language models. The question is clear: ``How can we ground abstract instructions into robust, feedback-driven execution in dynamic environments?'' As the scope of robotic applications expands, the core challenge has shifted: robots are no longer asked to repeat narrow, pre-programmed motions, but to understand flexible instructions and adapt to unpredictable situations. Existing methods have made progress on this front, typically following two strategies. The first learns behaviors directly from demonstrations, capturing motion trajectories with imitation or policy learning~\cite{coarsefineqatten, qattention, percieveractor}. The second relies on vision-language models (VLMs) to map natural language and visual input into actions~\cite{robomamba, openvla, rt2, rt1}. Both approaches have been effective in structured settings, but they falter when tasks involve open-vocabulary commands, new objects, or novel environments, where limited semantic understanding and adaptability restrict their use in real-world scenarios~\cite{imitationservey, vlachallenge}.

LLMs offer a promising path forward. They excel at reasoning and problem decomposition, and can translate high-level instructions into structured steps or even executable code~\cite{mathprompter, symbolicreasoning}. Frameworks that harness these capabilities demonstrate that LLMs can serve as powerful planners for robots~\cite{codeaspolicy, llmroboticsurvey, moka, rcecot}. However, these systems often operate in an open-loop manner: they generate plans once, without checking whether execution succeeds in practice. This makes them fragile in dynamic environments, where errors accumulate and hallucinations —plans that look valid in text but fail in the real world— can degrade performance~\cite{chainverification, robotmemoryintegrate,dadue}.

Recent research has taken steps toward closing the loop by integrating visual feedback for error detection and replanning~\cite{replanvlm, pchelintsev2025lerareplanningvisualfeedback, huang2022inner, skreta2024replan, Huang_2025}. Yet, most of these systems rely on a single, monolithic model, which creates bottlenecks when tasks are ambiguous or when reasoning and perception need to be specialized. Moreover, relying on unconstrained LLMs/VLMs raises safety concerns, as unchecked outputs can lead to unsafe or adversarial behaviors~\cite{zhang2025badrobot}.

\begin{figure}[h]
  \centering
  \includegraphics[width=1\textwidth]{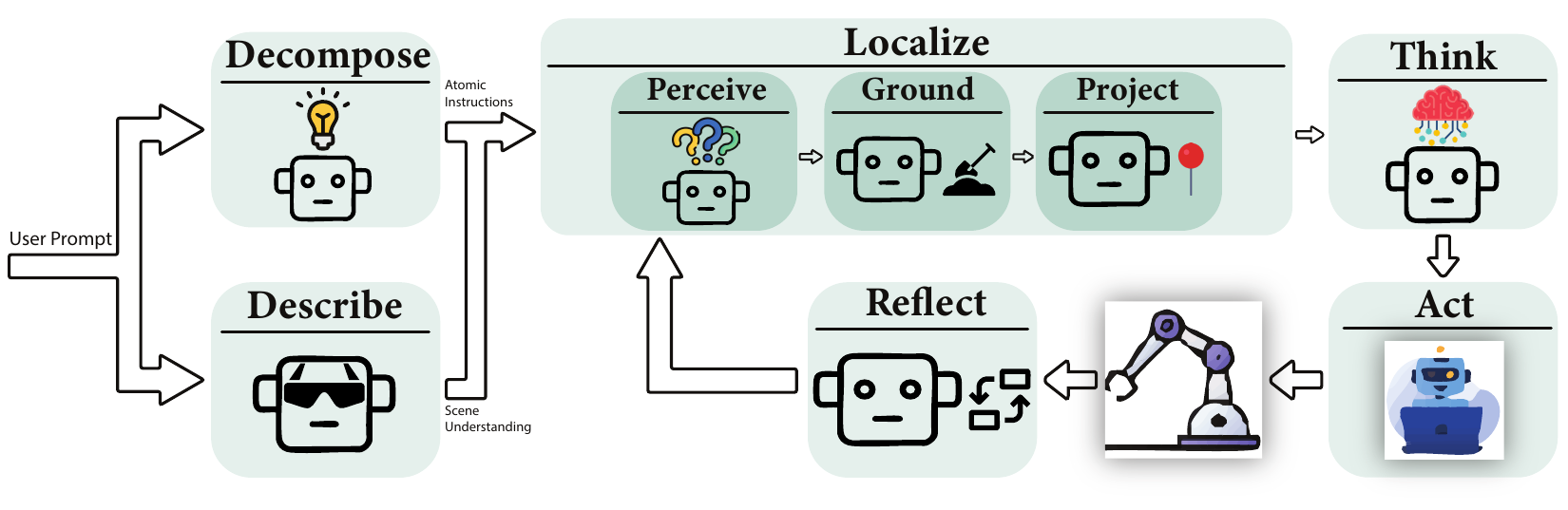}
  \caption{The MALLVi framework architecture. The pipeline processes user prompts through specialized agents: \textbf{Decompose} breaks instructions into atomic steps, \textbf{Describe} provides scene understanding, \textbf{Perceive} processes visual inputs, \textbf{Ground} localizes target objects, \textbf{Project} generates motion trajectories, \textbf{Think} coordinates high-level reasoning, \textbf{Act} executes robotic commands, and \textbf{Reflect} evaluates outcomes to enable iterative refinement and error recovery.}
  \label{fig:1}
\end{figure}

In this paper, we introduce MALLVi (Fig.~\ref{fig:1}), a multi-agent framework for robotic task planning that directly addresses these challenges. Instead of relying on monolithic models, MALLVi coordinates specialized agents for perception, planning, and reflection, enabling them to collaborate through a shared state. At its core, a decomposer agent translates human prompts into atomic instructions suitable for robotic execution. Subsequent agents handle environmental understanding, object localization, and trajectory planning through a low-level motion planner. A reflector agent continuously monitors the environment via visual feedback, providing a closed-loop that identifies and reactivates only the specific failing agent for efficient error recovery. This distributed, feedback-driven design enables MALLVi to disambiguate instructions, adapt to unexpected changes, and recover from errors—capabilities essential for real-world deployment. All source code and implementation details can be found at \href{https://github.com/ACTCL-SUT/mallvi}{this repository}.

Specifically, we:
\begin{itemize}
\item Propose MALLVi, a distributed framework that introduces a genuine multi-agent architecture for robotics, combining LLM-based planning with VLM-based monitoring in a self-correcting process.

\item Highlight the novel role of a reflector agent's targeted feedback loop, enabling reflection, error recovery, and adaptation through continuous environmental feedback.

\item Validate MALLVi in both simulation (VIMABench~\cite{jiang2023vima}, RLBench~\cite{james2020rlbench}) and real-world experiments, demonstrating substantial improvements in success rate across diverse manipulation tasks in a zero-shot setting.
\end{itemize}

\section{Related work}
\label{related_work}

\subsection{LLMs and VLMs for Robotic Task Planning}
The use of LLMs as high-level planners for robotics has grown rapidly in recent years. Frameworks such as \textit{Code-as-Policies}~\cite{codeaspolicy} treat LLMs as translators that convert natural language into parameterized API calls or executable code. \textit{Inner Monologue}~\cite{huang2022inner} was an early example of incorporating environmental feedback—including success and failure reports—to guide the LLM's planning. Subsequent work, such as LLM-Planner~\cite{song2023llm}, demonstrated few-shot planning for embodied agents by leveraging LLMs to generate sequences of pre-defined actions. Similarly, \textit{Tree-Planner}~\cite{hu2024treeplanner} iteratively constructs a task tree using an LLM, decomposing high-level goals into a sequence of executable subtasks.

A central challenge in these approaches is the open-loop nature of the initial plans. To mitigate this, recent research integrates visual feedback into replanning. For example, \textit{Replan}~\cite{skreta2024replan} combines an LLM for initial planning with a VLM to evaluate execution success, triggering replanning upon failure. \textit{ReplanVLM}~\cite{replanvlm} and \textit{LERa}~\cite{pchelintsev2025lerareplanningvisualfeedback} similarly use VLMs to detect visual errors and guide corrective action. \textit{Look Before You Leap}~\cite{Huang_2025} leverages GPT-4V to verify pre- and post-conditions for each planned step, while \textit{CoPAL}~\cite{copal} introduces a self-corrective planning paradigm where the LLM critiques and refines its own actions. Despite these advances, these systems are largely monolithic and limited in modularity, often handling planning, perception, and execution in a tightly coupled manner. They typically lack task-aware decomposition and flexible perception-action pipelines.

\subsection{Multi-Agent Collaboration and Reflection}
The concept of using multiple LLM agents to collaborate, debate, or critique each other has proven effective for complex problem-solving in non-embodied settings~\cite{multiagentproblem, wang-etal-2024-rolellm}. This idea is now being applied to robotics. \textit{RoCo}~\cite{roco} is a seminal work that introduces dialectic collaboration between multiple LLM-controlled robots for task planning.

Recent multi-agent approaches have further advanced collaboration in planning and manipulation. \textit{Wonderful Team}~\cite{wonderfulteam} presents a multi-agent VLLM framework where agents jointly generate action sequences from a visual scene and task description, integrating perception and planning in an end-to-end system. \textit{MALMM}~\cite{mlmm} employs three LLM agents (Planner, Coder, Supervisor) to perform zero-shot block and object manipulation tasks, incorporating real-time feedback and replanning to mitigate hallucinations and adapt to unseen tasks.

Building on prior multi-agent LLM frameworks, our work adopts a modular multi-agent approach for manipulation. Unlike previous methods, MALLVi tightly integrates perception, reasoning, and execution in a collaborative agent pipeline, enabling robust adaptation and closed-loop correction in complex environments.

\subsection{Open-Vocabulary Perception}
Robust manipulation requires not just object localization, but context-aware grounding to resolve ambiguities (e.g., ``the red block'' when there are multiple, referring to a past block). While foundational models like OWL-ViT~\cite{owlvit} and Grounding Dino~\cite{liu2023grounding} provide open-vocabulary detection, they lack situational context. Segmentation models such as the Segment Anything Model (SAM)~\cite{sam} offer general-purpose, high-quality segmentation masks across diverse object categories. These capabilities make them well-suited for downstream tasks such as grasp point extraction, where precise segmentation underpins reliable interaction. However, they do not incorporate contextual reasoning or task grounding. Approaches such as \textit{SayCan}~\cite{canisay} address this gap by incorporating environmental cues, and recent VLMs increasingly combine perception, grounding, and reasoning in a unified framework~\cite{qwenvl, pivot}. MALLVi builds on these advances by providing a modular, context-aware perception pipeline that integrates detection, segmentation, and grounding to enable precise manipulation.




\section{Methodology}
\label{Methodology}
The MALLVi framework implements a multi-agent, self-correcting pipeline for robotic manipulation. Given a high-level user instruction and a real-time image of the environment, the system hierarchically decomposes tasks into atomic subtasks, grounds each to visual inputs, plans execution trajectories, and adaptively refines actions based on feedback. 
Specialized agents communicate through object and memory tags, with automatic retry mechanism ensuring action success.

\subsection{Multi-Agent}

\begin{figure}[h]
  \centering
  \includegraphics[width=1.\textwidth]{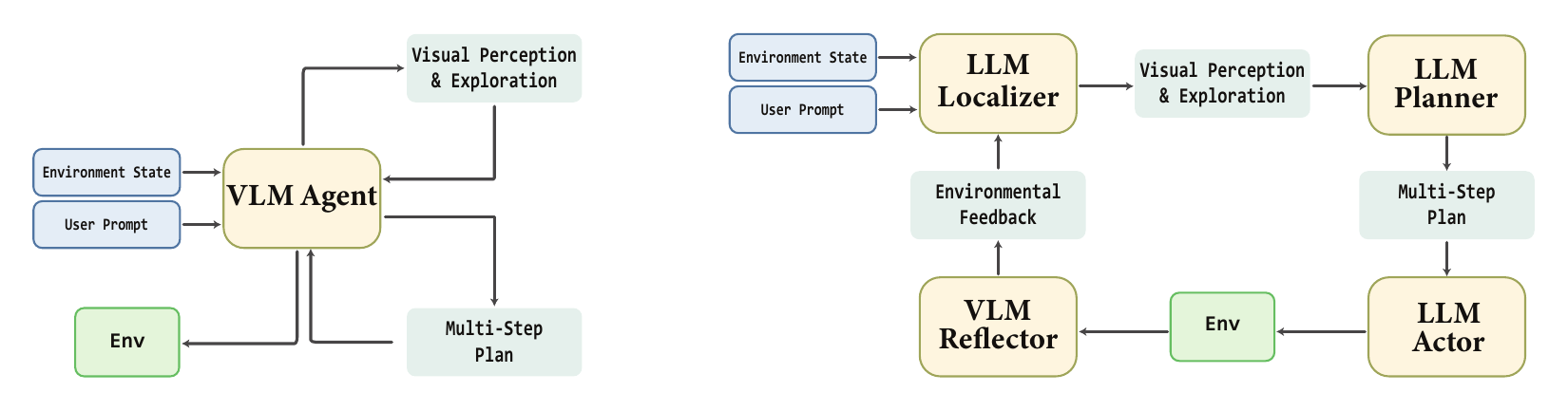} 
  \caption{Comparison between single-agent and multi-agent frameworks.}
  \label{fig:2}
\end{figure}

Single-agent frameworks often struggle with maintaining task focus and performing sequential reasoning in complex environments. As illustrated in Fig.~\ref{fig:2}, MALLVi mitigates these limitations by assigning specialized agents to distinct execution aspects. This modular design reduces hallucinations, supports iterative refinement through closed-loop feedback, and improves overall task execution by maintaining focus and coherence.

\subsubsection{Decomposer-Descriptor Agents}
\label{decomposer}
The \textbf{Decomposer} and \textbf{Descriptor} agents operate in parallel at the first stage of the MALLVi pipeline, providing complementary functions for task execution.

The Decomposer agent converts a high-level instruction into a structured sequence of atomic subtasks. Each subtask corresponds to a primitive action in the Actor agent’s vocabulary (e.g., \texttt{move}, \texttt{reach}, \texttt{push}) and is annotated with memory tags containing parameters such as object identities, positions, or contextual references. Subtasks are executed sequentially, with a retry mechanism that allows failed steps to be reattempted without replanning the entire sequence. Implementation details, including subtask representation, memory tagging, and fault-tolerant execution, are provided in Appendix section ~\ref{appendix decomposer}, and ~\ref{appendix descriptor}.

The Descriptor Agent generates a coarse representation of the environment using a vision-language model (VLM). It identifies objects, extracts the spatial relationships between them, and builds a spatial graph representing the scene. This graph enables the agent to reason about object configurations, constraints, and interactions, providing critical context for downstream perception, grounding, and planning agents.

By running in parallel, the Decomposer focuses on \textit{what} needs to be done (task decomposition), while the Descriptor focuses on \textit{where} and \textit{how} in the environment (scene representation and reasoning). Together, they align task objectives with environmental context from the outset.



\subsubsection{Localizer Agent}
\label{perceptor}
\begin{itemize}
    \item The \textbf{Perceptor} agent identifies task-relevant objects from the instruction and labels non-target objects. It refines grasping strategies (as explained in the Projector tool) across multiple attempts, adapting to subtask failures and improving manipulation precision.

    \item The \textbf{Grounder} agent localizes objects in the image plane by integrating outputs from multiple detectors (GroundingDINO and OwlV2) to ensure reliable detection even under partial failures. Beyond simple fusion, the agent employs a confidence-based selection mechanism: for each object, it weighs predictions from each detector according to model confidence and consistency with the spatial graph provided by the Descriptor agent. 
    This allows the Grounder to provide accurate bounding boxes for downstream planning. By combining multi-model detection and confidence weighting, the Grounder agent ensures robust, high-fidelity localization essential for manipulation in dynamic and unstructured environments.

    \item The \textbf{Projector} tool converts visual perception into actionable 3D grasp points, bridging the gap between scene understanding and robot execution.

    \textbf{Grasp Point Extraction:} Leveraging the Segment Anything Model (SAM), the agent identifies candidate grasp points on objects. Object-specific heuristics are applied to select appropriate points (e.g., edges for cylindrical objects, centers for rigid blocks). A verification step ensures that each grasp point lies within the object’s segmentation mask, enhancing reliability and precision.
    
    \textbf{3D Projection:} The extracted 2D grasp points are projected into 3D space using the depth map and the pinhole camera model. These 3D coordinates are subsequently converted into joint angles through inverse kinematics, producing executable targets for downstream planning and manipulation agents.
    
    By integrating grasp point extraction and 3D projection within a single module, the Projector agent provides a direct and reliable interface between visual perception and robotic action. This design enables precise and consistent generation of executable 3D targets, supports closed-loop feedback during task execution, and preserves the modularity of the MALLVi framework, allowing seamless interaction with downstream planning and manipulation agents.
\end{itemize}


\subsubsection{Thinker Agent}
\label{thinker}


The \textbf{Thinker} agent is an LLM responsible for translating high-level subtask information into actionable parameters for execution. It retrieves relevant objects (see section \ref{decomposer}) and determines 3D grasp points along with any required rotations. 

For tasks without prior memory (memoryless), the Thinker selects pick-and-place positions and rotations directly from the grasp points. For atomic instructions with associated memory tags, the agent identifies either source or target objects using the stored scene representation and spatial relationships, then computes corresponding pick-and-place positions and rotations based on the scene context.

\subsubsection{Actor Agent}
\label{actor}
The \textbf{Actor} agent executes the subtasks produced by the upstream agent. In both real-world deployments and benchmark scenarios, the Actor interfaces with the environment through a predefined API, receiving the action parameters from the Thinker and performing the corresponding manipulation. This modular design allows the Actor to remain agnostic to high-level reasoning or low-level motion planning while ensuring accurate execution of planned actions.

\subsubsection{Reflector Agent}
\label{reflector}
The \textbf{Reflector} agent is a Vision-Language Model (VLM) responsible for
verifying the execution of each subtask in real time. After the Actor executes a
subtask, the Reflector evaluates its success using visual feedback. Successfully
completed subtasks are removed from the execution queue, while failed subtasks
trigger a reattempt from the corresponding subtask. Beyond simple retries, the
Reflector produces natural-language explanations for failures, updates the
shared memory state, selectively reactivates the failing agent, and escalates to
full scene re-evaluation when necessary. In cases of repeated failures, the
Descriptor agent performs a complete scene re-analysis to update the visual
memory and account for changes such as object displacement. If the subtask
continues to fail after scene re-evaluation, the system infers an unrecoverable
failure and terminates execution with a structured failure report. As
demonstrated by the ablation results in Tables~\ref{tab:1}, \ref{tab:2}, and
\ref{tab:3}, this iterative, closed-loop verification and recovery mechanism is
essential for executing complex and sophisticated manipulation tasks. By
continuously monitoring task execution and applying bounded, targeted recovery,
the Reflector agent enhances reliability and generalization across diverse and
dynamic manipulation scenarios, while preserving the modular design of the
MALLVi framework.


\begin{figure}[h]
  \centering
  \includegraphics[width=1\textwidth]{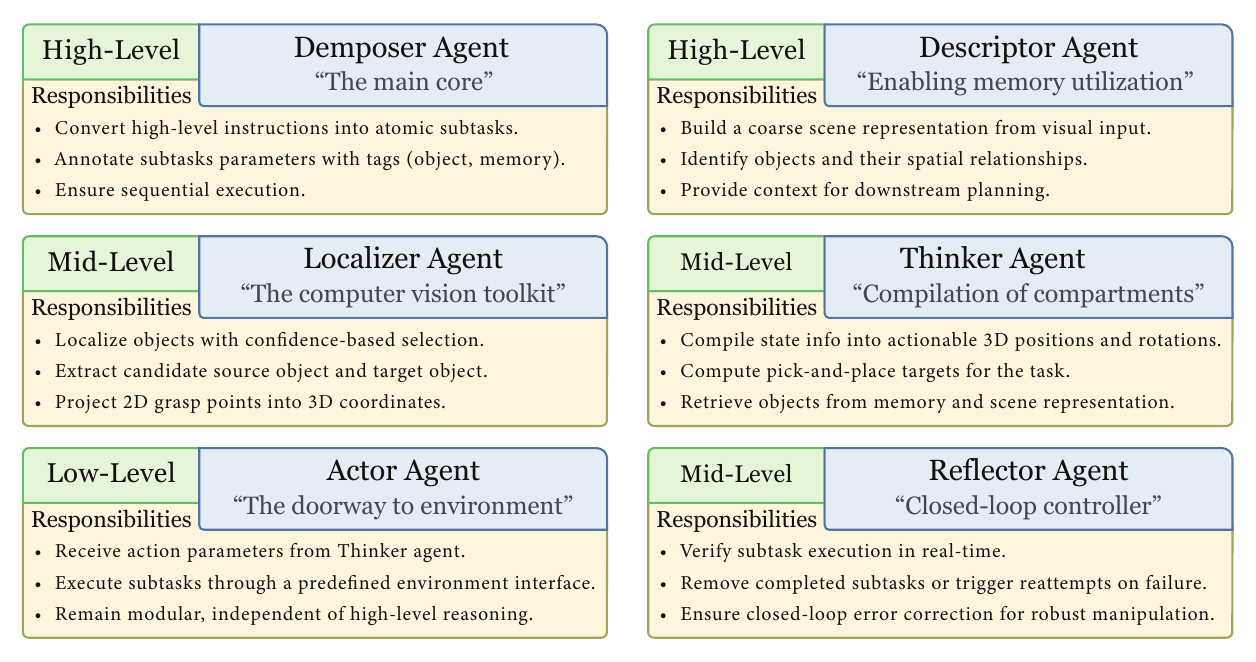} 
  \caption{Analysis of specialized agents and their roles in a multi-agent system. Each agent functions at a designated level (high, mid, or low) to address specific components of task execution, including instruction decomposition, memory utilization, object localization, task reasoning, action execution, and closed-loop feedback provision.}
  
  \label{fig:3}
\end{figure}

\section{Experiments and Results}
\label{experiments}

We evaluate MALLVi on both real-world manipulation tasks and benchmarked scenarios from VIMABench and RLBench. These tasks were selected for their alignment with real-world deployment settings, where the agent receives natural language instructions from users and perceives the environment solely through streaming camera input.

\textbf{Real-world tasks:} These are designed to reflect common robotic manipulation objectives:
\begin{itemize}
    \item \textbf{Place Food} -- tests accurate object placement.
    \item \textbf{Put Shape} -- evaluates shape-specific placement.
    \item \textbf{Stack Blocks} -- measures precision in stacking.
    \item \textbf{Shopping List} -- requires sequential task execution.
    \item \textbf{Put in Mug} -- tests fine-grained placement.
    \item \textbf{Math Ops} -- evaluates math reasoning.
    \item \textbf{Stack Cups} -- tests repetitive stacking skills.
    \item \textbf{Rearrange Objects} -- requires organizing multiple objects according to instructions.
\end{itemize}
Examples of task stages are shown in Fig.~\ref{fig:4}.

\begin{figure}[h]
  \centering
  \includegraphics[width=0.8\textwidth]{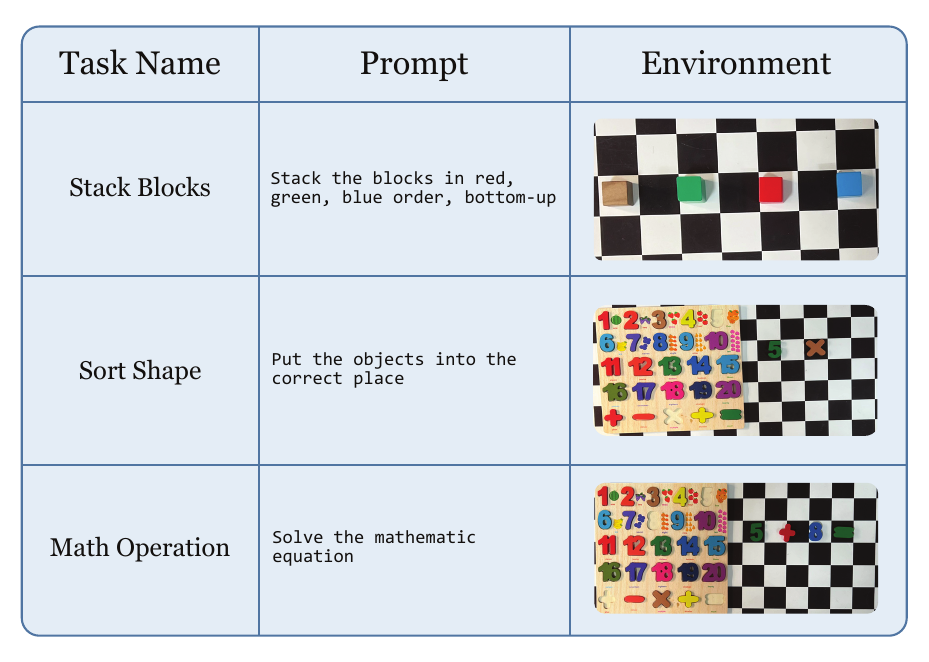} 
  \caption{Example of our real-world tasks. Stack Blocks, Sort Shape, and Math Operation each combine a specific prompt with a physical environment to assess an agent’s ability to act and solve problems in tangible settings.}
  \label{fig:4}
\end{figure}

\textbf{VIMABench tasks:} We selected a subset of VIMABench partitions with clear real-world analogues:
\begin{itemize}
    \item \textbf{Simple Manipulation} -- evaluates basic object handling.
    \item \textbf{Novel Concepts} -- tests the agent’s ability to generalize to unseen object--instruction combinations.
    \item \textbf{Visual Reasoning} -- requires reasoning under perceptual restrictions.
    \item \textbf{Visual Goal Reaching} -- measures scene understanding and goal-directed planning.
\end{itemize}

\textbf{RLBench tasks:} These tasks require diverse skill sets in simulated environments:
\begin{itemize}
    \item \textbf{Put in Safe} -- tests accurate object placement under safety constraints.
    \item \textbf{Put in Drawer} -- assesses sequential and goal-directed manipulation.
    \item \textbf{Stack Cups} -- evaluates repetitive stacking.
    \item \textbf{Place Cups} -- requires fine motor control in constrained settings.
    \item \textbf{Stack Blocks} -- measures precision and planning in multi-step tasks.
\end{itemize}

\subsection{Real-world tasks}
Real-world tasks capture conventional manipulation and reasoning skills. We reimplemented MALMM~\cite{mlmm}{VoxPoser~\cite{huang2023voxposer},ReKep~\cite{Huang2024ReKepSR} and used it as a baseline to evaluate our results in a real-world setting. As shown in Table~\ref{tab:1}, MALLVi achieves the
highest success rates across all tasks, outperforming prior methods, highlighting the benefit of closed-loop,
multi-agent planning in real-world settings.

\begin{figure}[h]
  \centering
  \includegraphics[width=1\textwidth]{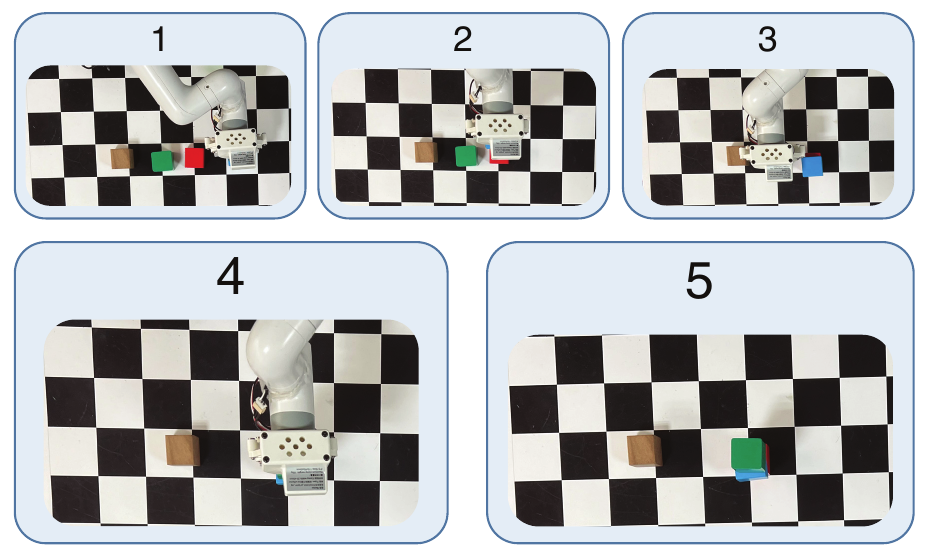} 
  \caption{A real-world example of the \textbf{Stack Blocks} task. MALLVi is asked to stack the blocks in the order red, blue and green. The wooden block acts as a distraction.}
  \label{fig:4}
\end{figure}
\begin{table}[htbp] 
  \centering
  \scriptsize
  \setlength{\tabcolsep}{2pt} 
  \begin{tabular}{lcccccccc}
    \toprule
    \textbf{Method} & \textbf{Place Food} & \textbf{Put Shape} & \textbf{Stack Blocks} & \textbf{Shopping List} & \textbf{Put in Mug} & \textbf{Math Ops} & \textbf{Stack Cups} & \textbf{Rearrange Objects} \\
    \midrule
    MALMM & 75 & 65 & 55 & 70 & 55 &  25 & 50 & -\\
    VoxPoser & 70 & 55 & 40 & 45 & 40 &  15 & 35 & 0\\
    ReKep & 80 & 85 & 75 & \textbf{90} & 75 & 60 & 40 & 60\\
    Single-Agent  & 25 & 10 & 15 & 10 & 30 & 5 & 10 & 0 \\
    w/o Reflector  & 85 & 60  & 60 & 65 & 55 & 70 & 50 & 45 \\
    \rowcolor{mallviColor} MALLVi (Ours) & \textbf{100} & \textbf{95} & \textbf{90} & \textbf{90} & \textbf{80} & \textbf{80} & \textbf{85} & \textbf{75} \\
    \bottomrule
  \end{tabular}
  \caption{Success rates (\%) on 8 real-world tasks with 20 repetitions.}
  \label{tab:1}
\end{table}

\subsection{VIMABench Tasks}

VIMABench tasks emphasize spatial reasoning, attribute binding, sequential planning, and state recall. Out of the 6 partitions and 17 tasks, we evaluated 12 tasks across 4 partitions suitable for our pipeline. As shown in Table~\ref{tab:2}, MALLVi outperforms prior methods, including \textit{Wonderful Team}~\cite{wonderfulteam}, \textit{CoTDiffusion}~\cite{10658623}, and \textit{PERIA}~\cite{NEURIPS2024_1f6af963}, achieving the highest success rates across most categories. Notably, the Reflector agent contributes to improvements in complex reasoning and goal-reaching tasks. For detailed results on each subtask, refer to Table~\ref{tab:appendix_vimabench_subtasks}.

\begin{table}[htbp] 
  \centering
  \scriptsize
  \setlength{\tabcolsep}{2pt} 
  \begin{tabular}{lcccccc} 
    \toprule
    \textbf{Method} & \textbf{Simple Manipulation} & \textbf{Novel Concepts}  & \textbf{Visual Reasoning} & \textbf{Visual Goal Reaching} \\
    \midrule
    Wonderful Team  & \textbf{100} & 85 & \textbf{90} & -  \\
    CoTDiffusion & 86 & 70 & 54 & 44 \\
    PERIA & 93 & 78 & 76 & 68 \\
    Single-Agent    & 25 & 10 & 15 & 10 \\
    w/o Reflector    & \textbf{100} & 80 & 30 & 40\\
    \rowcolor{mallviColor} MALLVi (Ours)      & \textbf{100} & \textbf{95} & \textbf{90} & \textbf{73} \\
    \bottomrule
  \end{tabular}
  \caption{Success rates (\%) on VIMABench tasks on 100 repetitions.}
  \label{tab:2}
\end{table}

\subsection{RLBench Tasks}

RLBench provides a large-scale benchmark for instruction-conditioned control. As a baseline, we compare MALLVi against PerAct~\cite{percieveractor}, which has been evaluated on similar tasks. Table~\ref{tab:3} shows that MALLVi consistently outperforms all baselines across the evaluated tasks, achieving the highest success rates in every category. The ablation without the Reflector agent demonstrates lower performance, highlighting the importance of closed-loop feedback for reliable execution. Single-agent performance remains limited, further emphasizing the benefits of multi-agent coordination in complex manipulation scenarios. Further details, including additional RLBench
tasks, more complex scenarios, and comparisons with MALMM and other ablations,
are provided in the Appendix~\ref{A.6} (Table~\ref{tab:5}).

\begin{table}[htbp] 
  \centering
  \scriptsize
  \setlength{\tabcolsep}{2pt} 
  \begin{tabular}{lccccccc} 
    \toprule
    \textbf{Method} & \textbf{Put in Safe}  & \textbf{Put in Drawer}  & \textbf{Stack Cups} & \textbf{Place Cups} & \textbf{Stack Blocks} \\
    \midrule
    PerAct & 44  & 68  & 0 & 0 & 36 \\
    Single-Agent          & 58  & 73  & 15 & 22 & 42 \\
    w/o Reflector         & 81  & 89  & 63 & 75 & 78 \\
    \rowcolor{mallviColor} MALLVi (Ours) & \textbf{92}  & \textbf{94}  & \textbf{83} & \textbf{96} & \textbf{90} \\
    \bottomrule
  \end{tabular}
  \caption{Success rates (\%) on RLBench tasks with 100 repetition.}
  
  \label{tab:3}
\end{table}

\subsection{Ablation Studies}
To better understand the contribution of individual components, we conduct the following ablations:

\subsubsection{Single-Agent Baseline} 
We collapse all functionality into a single LLM agent, removing explicit task decomposition and modular specialization. Tables~\ref{tab:1},\ref{tab:2}, and \ref{tab:3} compare this baseline with our multi-agent system. The results show that, while a single agent can handle simpler tasks, it struggles with compositional reasoning and grounding. In contrast, the multi-agent system leverages specialized agents, leading to higher accuracy and greater robustness.

\subsubsection{Without Reflector} 
We remove the Reflector agent, eliminating the retry mechanism. Although subtasks are still executed in sequence, no verification step is performed. Tables~\ref{tab:1},~\ref{tab:2}, and ~\ref{tab:3} compare the system with and without the Reflector. While the pipeline remains functional without it, verification and retry significantly improve reliability and overall task success rates. This is especially apparent for complex tasks, where potential for error is higher.

\subsubsection{Open-Source Substitution} 
We replace GPT-4.1-mini (the default MALLVi backbone LLM) with open source models, including Qwen~\cite{qwen2025qwen25technicalreport} + Qwen-VL~\cite{qwen2025qwen25technicalreport} (3B and 7B) and LLaMA 3~\cite{grattafiori2024llama3herdmodels} + LLaMA-Vision 3.2 (8B and 11B), to evaluate performance gaps between proprietary and publicly available systems. Table~\ref{tab:4} summarizes the results. Although open source models perform competitively on simple tasks, they underperform on compositional and multimodal tasks. However, because MALLVi separates the task into several smaller duties for each agent, the pipeline still generally demonstrates acceptable accuracy relative to model size, indicating a core strength of our approach.

\begin{table}[htbp] 
  \centering
  \scriptsize
  \setlength{\tabcolsep}{2pt} 
  \begin{tabular}{lcccc} 
    \toprule
    \textbf{Method} & \textbf{Simple Manipulation} & \textbf{Novel Concepts}  & \textbf{Visual Reasoning} & \textbf{Visual Goal Reaching} \\
    \midrule
    \textsc{GPT-4.1-Mini} \cite{} & 100 & 95 & 95 & 73  \\
    \textsc{Qwen-3B} w/ \textsc{Qwen-VL} & 70 & 54 & 10 & 46 \\
    \textsc{Qwen-7B} w/ \textsc{Qwen-VL} & 85 & 50 & 30 & 62 \\
    \textsc{LLaMA-3.1-8B} w/ \textsc{LLaMA-Vision-3.2-11B} & 80 & 50 & 27 & 59 \\
    \bottomrule
  \end{tabular}
  \caption{Success rates (\%) on open-source models over 100 repetitions.}
  \label{tab:4}
\end{table}

\section{Conclusion}
Our MALLVi framework leverages multiple LLM agents to plan and execute robotic manipulation tasks using closed-loop environmental feedback. Although this design enables robust high-level planning and iterative task refinement, it still relies on predefined atomic actions for execution, which constrains adaptability when the robot encounters unforeseen kinematic constraints, contact dynamics, or highly dynamic environments. This limitation reflects a broader trade-off between structured multi-agent reasoning and flexible low-level control.

Future work should explore the integration of adaptive execution mechanisms, such as reinforcement learning or imitation learning controllers, or differentiable motion planning modules. Such extensions would allow atomic actions to be adapted at deployment time, complementing the iterative reasoning and reflection already provided by the agents. In addition, incorporating more sophisticated perception and grounding modules could improve performance in tasks with novel objects, complex textures, or highly dynamic scenes.

MALLVi demonstrates that a multi-agent, closed-loop LLM framework can autonomously manage all key aspects of manipulation tasks, from perception and reasoning to high-level planning and reflection, leading to improved generalization and success rates. By combining structured reasoning with adaptive low-level execution, future iterations of MALLVi have the potential to achieve even greater robustness and autonomy in real-world robotic manipulation.
\bibliography{iclr2026_conference}
\bibliographystyle{iclr2026_conference}


\appendix
\section{Appendix}

\subsection{Agents}
Each agent serves a critical function in the end-to-end execution pipeline for user instructions. We demonstrate why each component is indispensable, accompanied by explaining its functionality in detail.

Our stack uses LangGraph\footnote{\href{https://www.langchain.com/langgraph}{link}}, enabling easy integration and changes, rendering ablation studies much simpler. A custom LangGraph wrapper with proper logging (for debugging purposes) was implemented. A log visualizer utilizing Dash\footnote{\href{https://dash.plotly.com/}{link}} serves as the primary debugging tool to visualize inter-agent interactions over time, using the log outputs from the LangGraph wrapper.
\\

\begin{lstlisting}[style=pythonICLR, caption={GraphState Class}]
class GraphState:
    
    taskname: str
    original_prompt: str
    initial_decomposition_done: bool
    decomposed_prompts: list[str]
    queue: list[str]
    current_prompt: str
    should_terminate: bool
    multi_object: bool

 
    object_of_interest: str
    not_object_of_interest: str
    all_objects: list[str]
    results: dict[str, dict]


    image: Image
    depth_image: Matrix
    camera_matrix: 3x3 Matrix
    rotation_matrix: 3x3 Matrix
    translation_vector: 3x1 Matrix


    grounder_output: list[Detection]
    grasp_points: list[GraspPoint2D]
    grasp_points_3d: list[GraspPoint3D]
    thinker_output: dict[str, ThinkerOutput]
    actor_output: dict[str, ActorOutput]
    reflection_output: dict[str, ReflectionResult]

    scene_description: Graph
    detected_objects: list[dict]
    descriptor_grasp_points: list[GraspPoint2D]
    descriptor_grasp_points_3d: list[GraspPoint3D]
\end{lstlisting}

\begin{lstlisting}[style=pythonICLR, caption={MultiAgentRoboticSystem Class}]
class MultiAgentRoboticSystem:

    def initialize_system(self) -> GraphState:
        state = GraphState()
        state.should_terminate = False
        state.initial_decomposition_done = False
        state.queue = []
        state.results = {}
        return state

    def run_main_pipeline(self, state: GraphState) -> GraphState:
        if not state.initial_decomposition_done:
            state = decomposer_node(state)
            state.initial_decomposition_done = True

        descriptor_result = descriptor_node(state)
      

        while state.queue and not state.should_terminate:
            state.current_prompt = state.queue.pop(0)

            perception_result = perceptor_node(state)

            grounding_result = grounder_node(state)

            segmentation_result = segmentor_node(state)

            projection_result = projector_node(state)

            planning_result = thinker_node(state)

            execution_result = actor_node(state)

            reflection_result = reflector_node(state)

            state.results[state.current_prompt] = {
                'thinker_output': state.thinker_output[state.current_prompt],
                'actor_output': state.actor_output[state.current_prompt],
                'reflection_output': state.reflection_output[state.current_prompt]
            }

        return state
\end{lstlisting}

\subsection{Decomposer as the main core}
\label{appendix decomposer}
The Decomposer agent is responsible for converting high-level instructions into structured, executable sequences of subtasks, providing the critical interface between abstract task specifications and the Actor agent’s primitive actions. This appendix details the internal mechanisms, representation, and execution logic of the Decomposer.

\subsubsection*{Subtask Generation}
Upon receiving a high-level instruction, the Decomposer generates a hierarchical sequence of subtasks. Each subtask corresponds to a primitive action in the Actor agent’s vocabulary, such as:
\begin{itemize}
\item \texttt{move} — navigating to a target location.
\item \texttt{reach} — extending an agent manipulator toward an object.
\item \texttt{push} — applying force to move an object.
\end{itemize}
Each subtask represents an atomic unit of work that can be executed independently while preserving the logical structure of the overall task.

\subsubsection*{Memory Tagging and Parameterization}
Subtasks are annotated with memory tags that provide all necessary execution parameters. These tags may include:
\begin{itemize}
\item Object identifiers and properties (e.g., size, type, affordances).
\item Spatial positions and orientations.
\item Contextual references derived from the environment or previous subtasks.
\end{itemize}
Memory tags enable the Actor agent to resolve ambiguities, maintain task consistency, and adapt dynamically if the environment changes during execution.

This agent's instruction prompt is shown in Fig.~\ref{fig:decomposer prompt 1}, and~\ref{fig:decomposer prompt 2} 

\subsection{Descriptor, enabling memory utilization}

\subsubsection*{Vision-Language Model (VLM) Integration}
\label{appendix descriptor}
The Descriptor leverages a pre-trained vision-language model to interpret raw sensory input and extract semantically meaningful information. Specifically, it:
\begin{itemize}
\item Detects and classifies objects in the environment.
\item Generates descriptive embeddings that capture object properties (e.g., type, color, size, affordances).
\item Associates textual and visual modalities, enabling grounding of high-level instructions to perceptual features.
\end{itemize}

\subsubsection*{Spatial Relationship Extraction}
Beyond individual object recognition, the Descriptor agent computes pairwise spatial relationships to capture the scene configuration. For each object pair, it encodes relationships such as:
\begin{itemize}
\item Relative positions (e.g., left, right, above, below).
\item Distances and proximities.
\item Interaction constraints (e.g., support, containment, adjacency).
\end{itemize}
These relational encodings are essential for reasoning about feasible actions, dependencies, and constraints in the environment.

\subsubsection*{Graph-Based Scene Representation}
The agent constructs a spatial graph where nodes correspond to detected objects and edges encode the extracted spatial relationships. This graph structure provides:
\begin{itemize}
\item A structured memory format for storing object and relational information.
\item An interface for downstream agents to query object configurations, constraints, and potential interactions.
\item Support for reasoning over both local neighborhoods (adjacent objects) and global scene layout.
\end{itemize}

\subsubsection*{Memory Utilization and Agent Interaction}
The spatial graph generated by the Descriptor agent is stored in a memory-accessible format, enabling other agents to:
\begin{itemize}
\item Query the environment efficiently without repeated perception.
\item Ground high-level instructions in the observed scene.
\item Plan and decompose tasks based on the current state and object interactions.
\end{itemize}
By serving as a centralized, structured memory representation, the Descriptor facilitates coordination among perception, planning, and execution agents. This agent's prompt is shown in Fig.~\ref{fig:descriptor prompt}


\subsection{Thinker compilation of compartments}
The Thinker agent functions as the reasoning and compilation module that converts high-level subtask information into actionable parameters for execution. It leverages the stored scene representation, memory tags, and spatial relationships to compile task-specific parameters required by the Actor agent. This agent's prompt is shown in Fig.~\ref{fig:thinker prompt 1}, and ~\ref{fig:thinker prompt 2}.

\subsubsection*{Parameterization of Subtasks}
The Thinker processes each subtask by:
\begin{itemize}
    \item \textbf{Contextual Analysis:} It examines the subtask description alongside the stored scene graph and memory tags to understand the objects, positions, and spatial constraints relevant to the task.
    \item \textbf{Action Parameter Computation:} Based on the context, the agent determines the parameters needed for execution. For example, for pick-and-place subtasks, it specifies the target positions, orientations, and rotations required to complete the action in accordance with the scene layout.
\end{itemize}

\subsubsection*{Handling Memoryless vs. Memory-Associated Tasks}
\begin{itemize}
    \item \textbf{Memoryless Tasks:} When no prior memory is associated, the Thinker collects pick-and-place parameters directly from the localizer agent's outputs, while inferring object orientations from the task’s description.
    \item \textbf{Memory-Associated Tasks:} For subtasks that reference prior memory, the Thinker uses the stored scene representation and relational information to identify source or target objects. It then determines the corresponding action parameters in the context of object positions and orientations.
\end{itemize}

\subsubsection*{Integration with Execution Agents}
The parameters generated by the Thinker are structured to interface directly with the Actor agent. Each parameterized subtask includes:
\begin{itemize}
    \item Target or involved objects (via memory references).
    \item Action-specific parameters such as positions and rotations.
    \item Any context or constraints derived from the scene representation.
\end{itemize}

\subsection{Necessity of Reflection}

Uncorrected actions can significantly increase task failure rates. Without such an agent, the manipulation pipeline effectively operates in an open-loop manner. During our experiments, we observed numerous instances where the robot failed to execute generated sub-tasks due to limited joint mobility and positional inaccuracies.
The VLM plays a crucial role by analyzing the scene and identifying faulty sub-tasks. This capability enables the system to reattempt execution, preventing what would otherwise be recorded as failure. Furthermore, the VLM can detect positional discrepancies between target objects and the end-effector, prompting the reasoning agent to revise pick-and-place coordinates accordingly. Details of the instructions provided to this agent can be observed in Fig.~\ref{fig:Reflector Prompt}.

\subsection{RLBench Experiment}
\label{A.6}
Table~\ref{tab:5} provides the full success-rate results across nine RLBench tasks evaluated over 100 repetitions. For completeness, we report the performance of our proposed MALLVi framework alongside three baselines: MALMM, the Single-Agent approach~\cite{article}, and our w/o Reflector ablation. As shown, MALLVi consistently achieves the highest success rates across the majority of tasks, highlighting its robustness and strong generalization across diverse manipulation skills
\begin{table}[ht]
  \centering
  \scriptsize
  \setlength{\tabcolsep}{2pt} 
  \begin{tabular}{lccccccccc} 
    \toprule
    \textbf{Method} & \shortstack{Basketball\\in Hoop}  & \shortstack{Close\\Jar}  & \shortstack{ Empty\\Container} & \shortstack{Insert\\in Peg} & \shortstack{Meat\\off Grill} & \shortstack{Open\\Bottle} & \shortstack{Put\\Block} & \shortstack{Rubbish\\in Bin} & \shortstack{Stack\\Blocks}\\
    \midrule
    MALMM & 82 & 76  & 59 & \textbf{67} & 88 & 91 & 93 & 81 & 47  \\
    Single-Agent & 45 & 37 & 34 & 26 & 41 & 78 & 89 & 41 & 22  \\
    w/o Reflector & 78 & 67 & 42 & 57 & 82 & 86 & 95 & 83 & 78  \\
    \rowcolor{mallviColor} MALLVi (Ours) & \textbf{89} & \textbf{81} & \textbf{71} & 66 & \textbf{94} & \textbf{93} & \textbf{100} & \textbf{91} & \textbf{90}  \\
    \bottomrule
  \end{tabular}
  \caption{Success rates (\%) on RLBench tasks with 100 repetition.}
  \label{tab:5}
\end{table}

\subsection{VIMABench Experimental Details}
VIMABench consists of 17 tabletop scenarios in an OpenAI Gym environment, with objects of 29 shapes, 17 colors, and 65 textures. The manipulation tasks range from simple manipulation to novel concepts and visual reasoning. An example of VIMABench execution frames and its multimodal prompts can be observed in Fig.~\ref{fig:vima}.

\begin{figure}[h]
    \centering
    \includegraphics[width=0.9\linewidth]{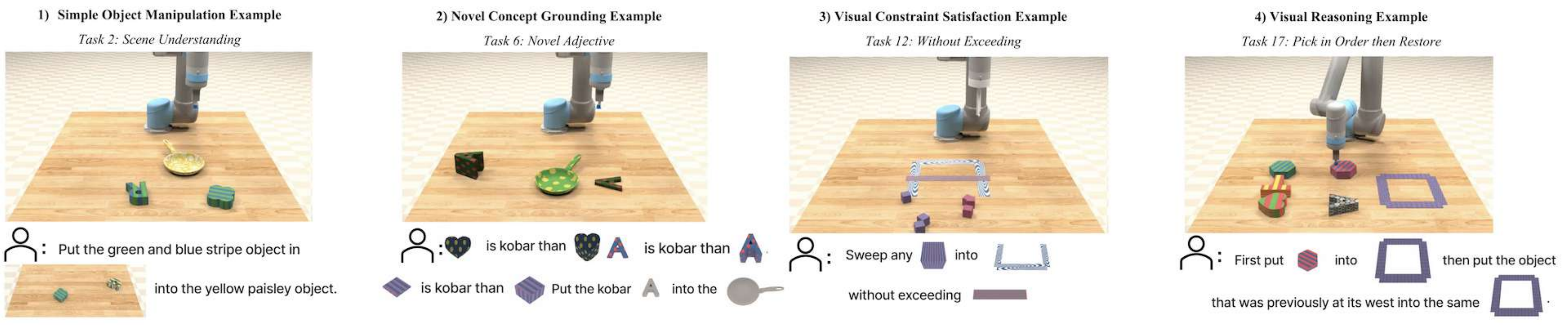}
    \caption{Credits to \cite{wonderfulteam} for the figure. Prompts for \textit{Visual Manipulation}, \textit{Novel Nouns}, \textit{Without Exceeding}, and \textit{Pick in Order then Restore} tasks.}
    \label{fig:vima}
\end{figure}
  
\begin{table}[htbp]
  \centering
  \scriptsize
  \setlength{\tabcolsep}{4pt}
  \begin{tabular}{llc}
    \toprule
    \textbf{Category} & \textbf{Subtask} & \textbf{MALLVi (Ours)} \\
    \midrule
    \multirow{3}{*}{Simple Manipulation} 
      & visual manipulation & 100 \\
      & scene understanding & 100 \\
      & rotate & 100 \\
    \midrule
    \multirow{3}{*}{Novel Concepts} 
      & novel adj & 95 \\
      & novel noun & 100 \\
      & novel adj and noun & 90 \\
    \midrule
    \multirow{4}{*}{Visual Reasoning} 
      & same texture & 95 \\
      & same shape & 90 \\
      & manipulate old neighbor & 90 \\
      & follow order & 85 \\
    \midrule
    \multirow{2}{*}{Visual Goal Reaching} 
      & rearrange &  70 \\
      & rearrange then restore & 76 \\
    \bottomrule
  \end{tabular}
  \caption{Detailed success rates (\%) for all VIMABench subtasks. This table provides a breakdown of the aggregated scores shown in Table \ref{tab:2}.}
  \label{tab:appendix_vimabench_subtasks}
\end{table}

\subsection{Optimal Grasp Point}
A robot's end-effector requires precise coordinates for stable grasping. While bounding boxes provide object localization, simply using their center point as the grasp position proves suboptimal for many objects - particularly those with irregular shapes, surface holes, or non-uniform geometry.

To address this limitation, we employ the Segment Anything Model (SAM)~\cite{sam} to generate precise segmentation masks for all detected objects. These binary masks accurately delineate object boundaries while excluding void regions. the objects are first categories to four classes based on their geometry : round perfect objects, rimmed ones, ... and irregular shapes. for the first three categories the grasping point is assumed manually for the latter, we compute an optimal grasp position using:

\begin{equation}
r^* = \min \{r \mid \text{mask}[C_x + r\cos\theta, C_y + r\sin\theta] = 1 \}
\end{equation}

where \((C_x, C_y)\) denotes the object's centroid, calculated as the mean position of all mask pixels, \(\theta \sim \mathcal{U}(0, 2\pi)\) is a uniformly distributed random angle, and \(r^*\) represents the minimal radial distance from centroid to mask boundary along direction \(\theta\). Refer to Fig.~\ref{fig:banana} for an example of grasping point calculation.

\begin{figure}[h]
    \centering
    \includegraphics[width=\linewidth]{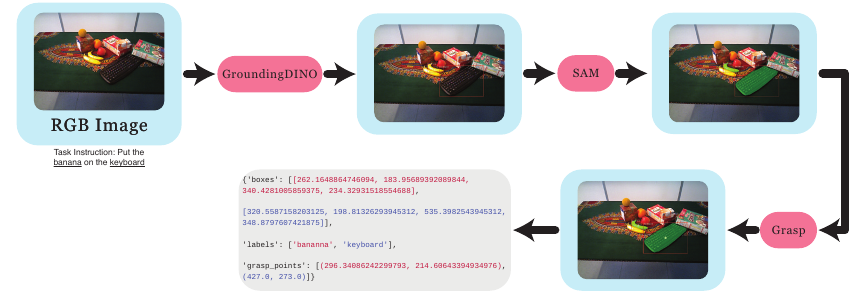}
    \caption{The pipeline is instructed to put the banana on the keyboard. The optimal grasp point for the robot's end-effector is thus determined.}
    \label{fig:banana}
\end{figure}

\subsection*{Real world Task Descriptions}

Below we describe the tasks used in our real world evaluation. Each task corresponds to a realistic robotic manipulation or reasoning scenario, designed to test grounding, planning, and execution capabilities.

\subsubsection*{Place Food}
The robot is given food items (e.g., an apple or banana) and instructed to place them in a designated location such as a plate or bowl. 
This task evaluates the agent’s ability to recognize semantic categories (food vs. non-food), perform spatial placement, and follow commonsense constraints (e.g., food should not be placed in inappropriate containers like shoes).

\subsubsection*{Put Shape in Math Sorter}
In this task, the traditional shape-sorting toy is adapted for symbolic reasoning. 
Instead of geometric shapes (circle, square, triangle), the cutouts correspond to \textbf{numbers or mathematical operators} (e.g., ``3'', ``7'', ``+'', ``-''). 
The robot must pick the correct block, recognize its symbolic label, and insert it into the matching slot. 

This task tests the integration of \textbf{symbol recognition} and \textbf{physical manipulation}. 
The robot must not only align the block physically to fit the slot but also correctly ground the abstract symbol (distinguishing, for example, between a number and an operator). 
Errors may occur from visual misclassification of symbols, confusion between similar digits, or incorrect orientation during insertion.

\subsubsection*{Stack Blocks}
The robot must stack a set of cubic blocks on top of each other to form a stable tower. 
This requires sequential action planning, stability estimation (avoiding imbalance), and careful execution. 
Failures typically arise from slippage or misalignment during stacking, making this a robust test of dexterity and control.

\subsubsection*{Shopping List}
The robot is given a shopping list (e.g., ``apple, orange, and milk'') and must retrieve the specified items from a set of available objects while ignoring distractors. 
This task evaluates \textbf{multi-step reasoning}, \textbf{object recognition}, and \textbf{memory maintenance} (keeping track of which items have already been collected).

\subsubsection*{Put Object in Mug}
The robot is asked to place a small object (e.g., spoon, pen, or sugar packet) inside a mug. 
This requires spatial reasoning about container affordances and careful placement to avoid dropping or misaligning the object. 
The task is representative of daily-life kitchen or office manipulations.

\subsubsection*{Math Operation}
In this task, arithmetic reasoning is combined with physical object manipulation. 
The robot is presented with blocks representing numbers (e.g., a block labeled ``9'' and a block labeled ``4''). 
The instruction specifies a math operation such as \textit{``place the result of 9 plus 4''}. 
To complete the task, the robot must:

\begin{enumerate}
    \item \textbf{Interpret the instruction:} Identify the operands and operation (e.g., $9 + 4$).
    \item \textbf{Compute the result symbolically:} Perform the arithmetic ($9 + 4 = 13$).
    \item \textbf{Ground the result physically:} Locate the correct answer block (``13'') from a set of candidate number blocks scattered in the workspace.
    \item \textbf{Manipulate the block:} Pick up the correct block and place it in the designated answer area.
\end{enumerate}

This task evaluates the integration of \textbf{symbolic reasoning} (arithmetic computation) and \textbf{embodied action} (locating and manipulating the correct block). 
Errors may arise from miscalculating the arithmetic operation, failing to ground the symbolic answer in the physical workspace, or incorrectly manipulating the chosen block.

\subsubsection*{Stack Cups}
The robot must stack a set of cups in a nested manner (placing one inside another) or create a vertical tower (placing them upright). 
The task requires reasoning about \textbf{object affordances}, \textbf{hollow geometry}, and \textbf{symmetry constraints}. 
Errors often occur if the robot fails to align the cup’s opening correctly.

\subsubsection*{Rearrange Objects}
The robot is tasked with rearranging a set of objects from one spatial configuration to another (e.g., ``place the book to the left of the laptop, and move the pen to the right of the notebook''). 
This task stresses \textbf{relative spatial reasoning}, \textbf{planning multiple sequential moves}, and avoiding collisions while repositioning objects.

\subsection{Computer Vision}
\label{calibration}

The grounders produce a bounding box in pixel space, denoted by the 2D point 
\(\mathbf{c} = \begin{bmatrix}u & v\end{bmatrix}^\intercal\). However, to solve the inverse kinematics 
problem for robotic manipulation, the end-effector requires target coordinates in real-world 3D space, 
represented by \(\mathbf{r} = \begin{bmatrix}X & Y & Z \end{bmatrix}^\intercal\). Therefore, a transformation 
\(T\) is required to convert pixel-space coordinates into real-world coordinates, such that 
\(T(u, v) = (X, Y, Z)\).

According to the pinhole camera model~\cite{Harltey2006MultipleVG}, the inverse mapping from real-world 
coordinates to pixel-space coordinates, denoted by \(T^{-1}(X, Y, Z) = (u, v)\), is described by the 
following projection equation:

\begin{equation}
z_{axial} \begin{bmatrix}
u \\ v \\ 1
\end{bmatrix}
=
\mathbf{K}
\begin{bmatrix}
\mathbf{R} \mid \mathbf{t}
\end{bmatrix}
\begin{bmatrix}
X \\ Y \\ Z \\ 1
\end{bmatrix}
\end{equation}

In this formulation, \(\mathbf{K}\) is the intrinsic camera matrix that describes the internal characteristics 
of the camera. It is given by:

\[
\mathbf{K} = \begin{bmatrix}
f_u & \alpha & u_0 \\
0 & f_v & v_0 \\
0 & 0 & 1
\end{bmatrix}
\]

The matrix \(\mathbf{R}\) represents the \(3 \times 3\) rotation from the real-world coordinate frame to the 
camera coordinate frame. The vector \(\mathbf{t}\) represents the \(3 \times 1\) translation from the real-world 
origin to the camera origin. The scalar \(z_{axial}\) is a scaling factor that accounts for the axial distance from the real-world point to the camera's principle point.

To obtain the \(z_{axial}\) value, we resort to stereo vision. Stereo vision consists of two cameras with identical configurations fixed at a predetermined horizontal distance apart, both looking towards the scene. In such a setup, \(z_{axial}\) can be derived as:

\begin{equation}
z_{axial} = \frac{B\:f_c}{d}
\end{equation}

Where \(B\) represents the baseline distance between the two cameras (measured in meters), \(f\) represents the focal length (in pixels), and \(d\) is the disparity (measured in pixels) between the projections of the same 3D point in both images.

To recover the real-world position \(\mathbf{r}\) from a pixel-space point \((u, v)\), the following equation is used:

\begin{equation}
\mathbf{r} =
\begin{bmatrix}
X \\ Y \\ Z
\end{bmatrix}
=
\mathbf{R}^{-1}
\left(
z_\text{{axial}} \mathbf{K}^{-1}
\begin{bmatrix}
u \\ v \\ 1
\end{bmatrix}
-
\mathbf{t}
\right)
\end{equation}

The intrinsic matrix \(\mathbf{K}\), the rotation matrix \(\mathbf{R}\), and the translation vector \(\mathbf{t}\) 
are obtained through standard camera calibration procedures.

\subsection{Prompts}
The complete instruction prompts to language models are provided in Figures \ref{fig:decomposer prompt 1} through \ref{fig:Reflector Prompt}
\begin{figure}[h]
    \centering
    \includegraphics[width=\linewidth]{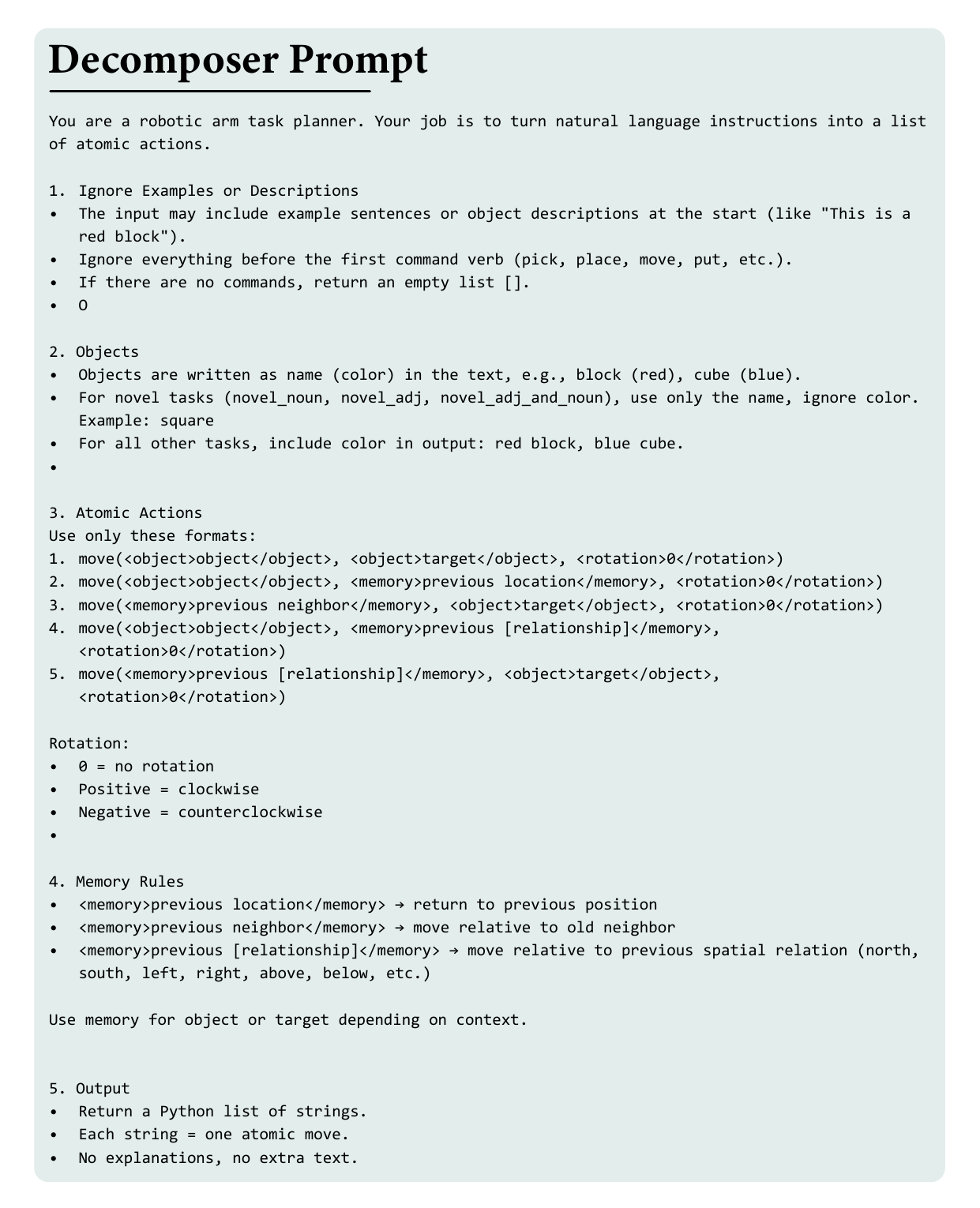}
    \caption{Decomposer prompt}
    \label{fig:decomposer prompt 1}
\end{figure}

\begin{figure}[h]
    \centering
    \includegraphics[width=\linewidth]{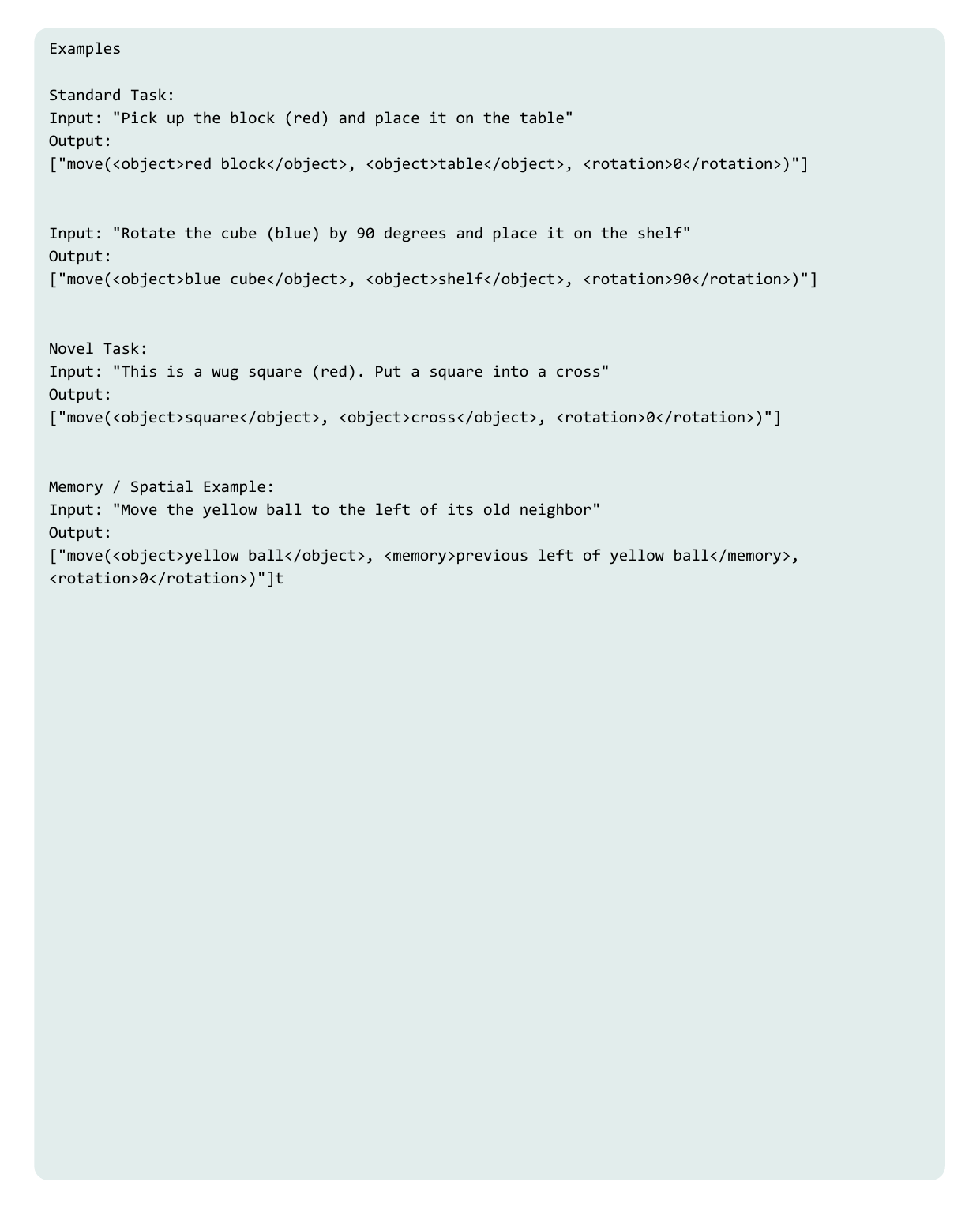}
    \caption{Decomposer prompt}
    \label{fig:decomposer prompt 2}
\end{figure}
\begin{figure}[h]
    \centering
    \includegraphics[width=\linewidth]{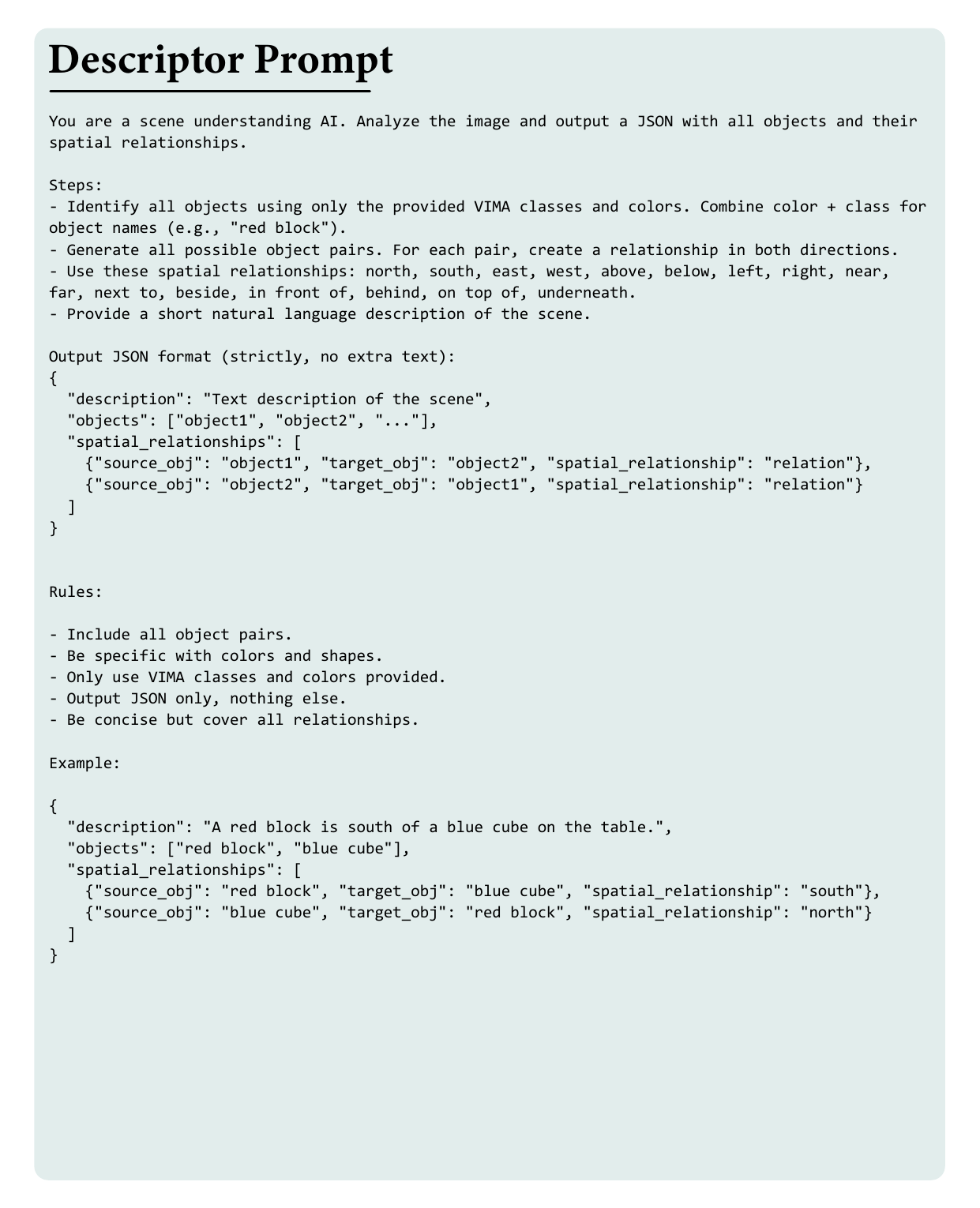}
    \caption{Descriptor prompt}
    \label{fig:descriptor prompt}
\end{figure}

\begin{figure}[h]
    \centering
    \includegraphics[width=\linewidth]{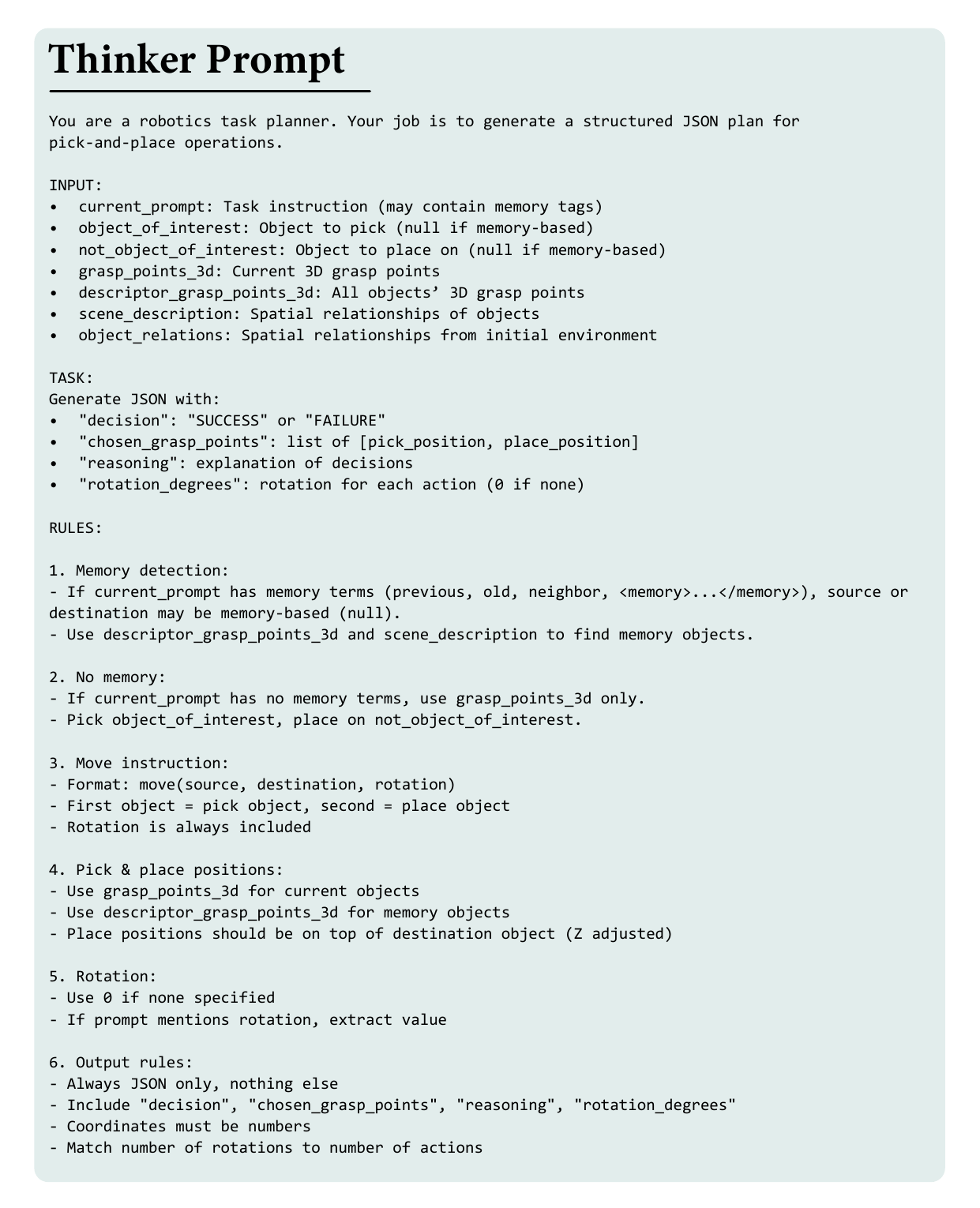}
    \caption{Thinker prompt}
    \label{fig:thinker prompt 1}
\end{figure}
\begin{figure}[h]
    \centering
    \includegraphics[width=\linewidth]{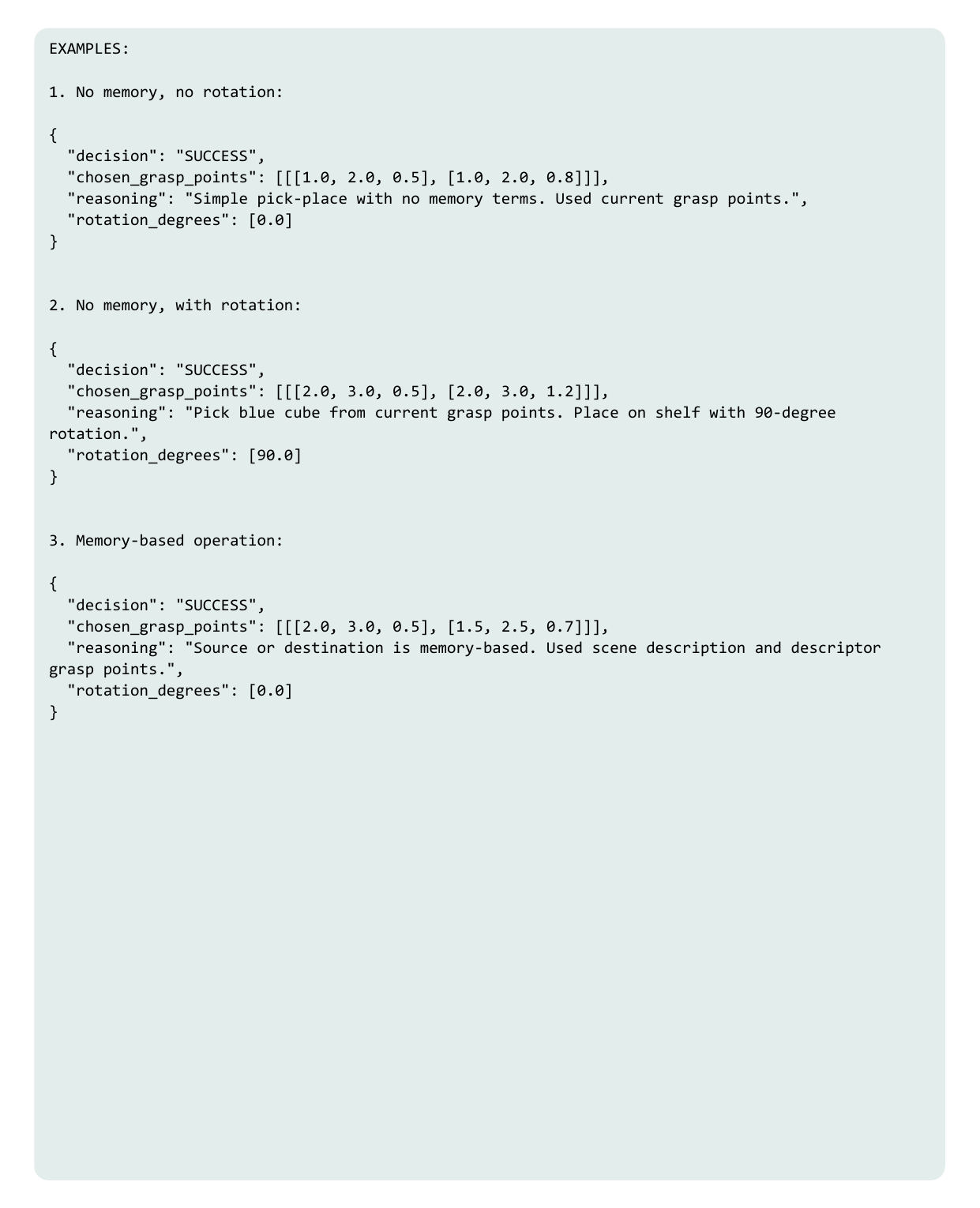}
    \caption{Thinker prompt}
    \label{fig:thinker prompt 2}
\end{figure}
\begin{figure}[h]
    \centering
    \includegraphics[width=\linewidth]{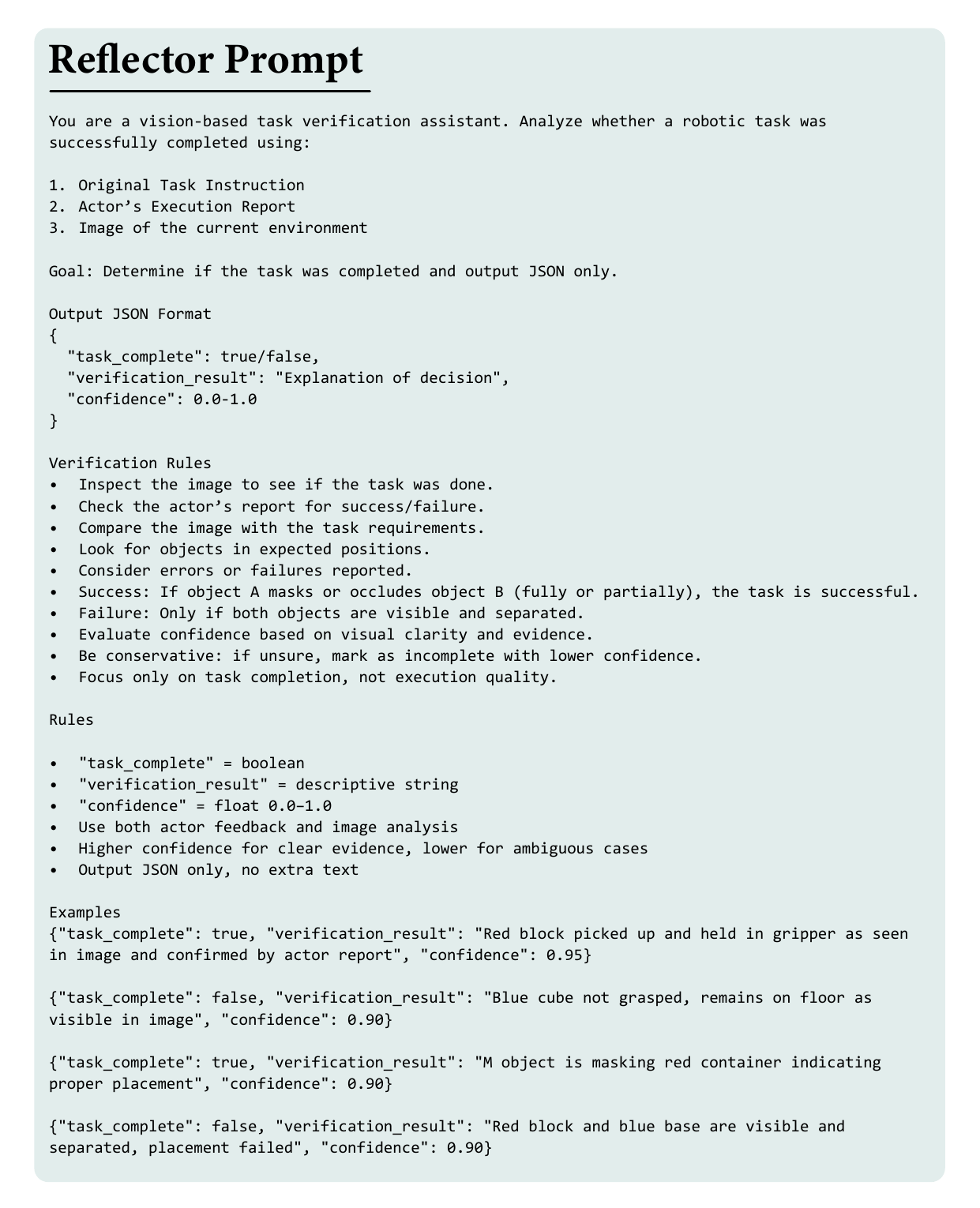}
    \caption{Reflector prompt}
    \label{fig:Reflector Prompt}
\end{figure}

\end{document}